\documentclass[lettersize,journal]{IEEEtran}
\usepackage{amsmath,amsfonts}
\usepackage{algorithmic}
\usepackage{algorithm}
\usepackage{array}
\usepackage[caption=false,font=normalsize,labelfont=sf,textfont=sf]{subfig}
\usepackage{textcomp}
\usepackage{stfloats}
\usepackage{url}
\usepackage{verbatim}
\usepackage{graphicx}
\usepackage{cite}
\usepackage{booktabs}
\usepackage{enumitem}
\usepackage{svg}

\usepackage{xcolor}
\usepackage{xpatch}

\begin{document}

\title{RASLF: Representation-Aware State Space Model for Light Field Super-Resolution}

\author{Zeqiang Wei, Kai Jin, Kuan Song, Xiuzhuang Zhou, Wenlong Chen, Min Xu\textsuperscript{*}
\noindent 
\thanks{\hspace{-1em}Corresponding author: Min Xu.}
\thanks{\hspace{-1em}Zeqiang Wei, Min Xu, and Wenlong Chen are with the Capital Normal University Information Engineering College, Beijing 100048, China (email: weizeqiang@cnu.edu.cn, xumin@cnu.edu.cn, chenwenlong@cnu.edu.cn).
Kai Jin is with the Bigo Technology Pte. Ltd., Beijing 100020, China (email: jinkai@bigo.sg). 
Kuan Song is with the Explorer Global (Suzhou) Artificial Intelligence Technology Co., Ltd., Suzhou, Jiangsu 215123, China (email: songkuan@explorer.global).
Xiuzhuang Zhou are with Beijing University of Posts and Telecommunications, Beijing 100088, China (email: xiuzhuang.zhou@bupt.edu.cn).
}
}


\maketitle

\begin{abstract}

Current SSM-based light field super-resolution (LFSR) methods often fail to fully leverage the complementarity among various LF representations, leading to the loss of fine textures and geometric misalignments across views.
To address these issues, we propose RASLF, a representation-aware state-space framework that explicitly models structural correlations across multiple LF representations.
Specifically, a Progressive Geometric Refinement (PGR) block is created that uses a panoramic epipolar representation to explicitly encode multi-view parallax differences, thereby enabling integration across different LF representations.
Furthermore, we introduce a Representation-Aware Asymmetric Scanning (RAAS) mechanism that dynamically adjusts scanning paths based on the physical properties of different representation spaces, optimizing the balance between performance and efficiency through path pruning.
Additionally, a Dual-Anchor Aggregation (DAA) module improves hierarchical feature flow, reducing redundant deep-layer features and prioritizing important reconstruction information.
Experiments on various public benchmarks show that RASLF achieves the highest reconstruction accuracy while remaining highly computationally efficient.

\end{abstract}


\section{Introduction}

\IEEEPARstart{L}{ight} field (LF) imaging captures both the spatial intensity and angular information of light rays, offering rich geometric priors that facilitate various downstream applications, including depth estimation \cite{cui2024NonLambertian, chao2024occcasnet} and refocusing \cite{yang2023joint, ban2024focus}.
However, the physical constraints of imaging sensors create an inherent trade-off between spatial and angular resolutions, usually resulting in sub-aperture images (SAIs) with limited spatial detail.
Therefore, light field super-resolution (LFSR) focuses on reconstructing high-quality details from low-resolution data, with the main challenge being to recover high-frequency textures while maintaining strict geometric consistency across views within the inherent spatial-angular structure.

Recently, SSMs \cite{chen2013state} have been introduced into LFSR to leverage linear computational complexity and model long-range dependencies, demonstrating promising effectiveness.
Unlike CNNs, limited by local receptive fields, and Transformers, burdened by quadratic complexity, SSMs are theoretically better at capturing long-range spatial-angular correlations in high-dimensional LF data.

Despite these advantages, current SSM-based LFSR methods \cite{Gao_2024_ACCV, xia2024, wei2025l2fmamba} still find it challenging to effectively capitalize on the diverse representations of LF data.
Specifically, many approaches limit their focus to a single LF domain, thereby overlooking the structural complementarity offered by other representations.
Even for methods that aim to incorporate multiple representations, the lack of a robust modeling framework often leads to heuristic aggregation that doesn't explicitly capture complex cross-view dependencies and long-range spatial-angular correlations.

Motivated by recent findings \cite{Pei_Huang_Xu_2025, 10.1609/aaai.v39i8.32934, Guo_2025_CVPR} that multi-directional scanning in image-based SSMs often induces feature redundancy, we observe that existing SSM-based LFSR methods typically adopt a representation-agnostic configuration that uses a uniform set of scanning paths for all LF representations.
Such designs ignore the inherent structural differences among LF representations. 
For instance, spatial-angular textures usually exhibit more balanced dependencies across different directions and therefore benefit from multi-directional scanning, whereas epipolar lines follow clear directional trajectories, making some scanning paths unnecessary.
Therefore, a uniform scanning strategy causes unnecessary computational overhead and reduces feature focus in highly structured LF representations.

To overcome these limitations, we introduce RASLF, a representation-aware state-space framework designed for LFSR. 
Specifically, we developed a Progressive Geometric Refinement (PGR) block that leverages a Panoramic Epipolar Representation to transform fragmented observations into a globally coherent geometric space, facilitating explicit parallax-aware feature interaction.
Building on this, a Representation-Aware Asymmetric Scanning (RAAS) strategy is introduced to align sequential modeling trajectories with the structural characteristics of different LF representations, thereby reducing computational redundancy while reinforcing geometric constraints.
Moreover, a Dual-Anchor Aggregation (DAA) module is incorporated to regulate hierarchical feature propagation, thereby filtering deep-layer redundancy and focusing computational budget on reconstruction-critical cues. 
Extensive evaluations confirmed that RASLF offers a better balance between reconstruction accuracy and computational efficiency across multiple benchmarks.

In summary, our primary contributions are as follows:
\begin{enumerate}
    \item We proposed a Progressive Geometric Refinement (PGR) block and a Panoramic Epipolar Representation, which jointly transform fragmented local constraints into a globally coherent geometric structure, significantly enhancing cross-view consistency.
    \item We designed a Representation-Aware Asymmetric Scanning (RAAS) strategy that aligns sequential modeling paths with the physical and structural characteristics of diverse LF representations, thereby reducing redundancy and computational overhead.
    
    \item To further improve feature utilization, we designed a Dual-Anchor Aggregation (DAA) module to optimize hierarchical feature propagation and suppress redundancy along the network hierarchy.

    \item The proposed RASLF achieves a state-of-the-art (SOTA) balance between reconstruction quality and inference efficiency, as validated by extensive experiments on public LF datasets.
\end{enumerate}

\section{Related Work}

\subsection{Light Field Representations}
Light field data is typically parameterized as a 4D function $L(u, v, x, y)$, following the two-plane parameterization \cite{Levoy1996}. 
Here, $(u, v)$ denotes the angular coordinates of the camera array, and $(x, y)$ represents the spatial coordinates within each view. 
This high-dimensional structure captures both spatial textures and angular correlations, which can be broken down into three functionally distinct 2D representations:

\begin{itemize}[leftmargin=*,itemindent=0pt]
    \item Sub-Aperture Images (SAI): By fixing the angular coordinates at $(u, v)$, an SAI $I_{u,v}(x, y)$ is obtained. SAIs resemble traditional 2D images and are primarily used to extract spatial features. 

    \item Macro-Pixel Images (MacPI): A MacPI $I_{x,y}(u, v)$ is formed by gathering pixels from all viewpoints at a fixed spatial location $(x, y)$. It encapsulates the angular distribution of light rays. 

    \item Epipolar Plane Images (EPI): By fixing one spatial and one angular dimension, the EPI $I_{y,v}(x, u)$ or $I_{x, u}(y, v)$ is generated. 
Since the parallax of a scene point is proportional to its depth, scene objects appear as directional linear structures with different slopes in EPIs \cite{Bolles1987}. 
This structural anisotropy defines EPIs as the primary domain for enforcing geometric consistency.
\end{itemize}

In the LFSR task, these three representations provide different but complementary viewpoints.
Early methods \cite{LFCNN2017, LFNet2018} predominantly performed spatial super-resolution on independent SAIs, a practice that treats the light field as a set of isolated 2D images while ignoring angular consistency.
To establish cross-dimensional correlations, spatial-angular interaction paradigms were developed to model the relationship between spatial textures and angular distributions by synergistically leveraging SAIs and MacPIs.
To further ensure geometric consistency, subsequent research introduced explicit EPI-based constraints on parallax slopes, yielding multi-representation frameworks that integrate spatial, angular, and epipolar information.

\subsection{Methods based on Spatial-Angular Interaction}
Spatial-angular interaction methods circumvent the high complexity of direct 4D processing by decomposing feature extraction into dimensionally decoupled 2D operations.
LF-InterNet \cite{Wang2020inter} utilizes parallel branches to iteratively exchange spatial and angular information. 
To mitigate parallax-induced misalignments inherent in decoupled paradigms, LF-DFnet \cite{Wang2021dfnet} introduced the Angular Deformable Alignment Module (ADAM) for non-rigid feature warping, while LF-IINet  \cite{cong2023lfdet}  refined interactions via parallel intra-inter view branches.
HDDRNet \cite{Meng2021HDDRNet} further simulated 4D correlations by using dense residual connections between the SAI and MacPI representations to enable intensive feature reuse.
However, the localized receptive fields of CNNs preclude the capture of long-range dependencies across distant viewpoints or large spatial structures.

To overcome this, transformer-based architectures utilize self-attention mechanisms to aggregate global context. 
LFT \cite{liang2022light} implements this through alternating spatial and angular Transformer modules, while DPT \cite{wang2022detail} and M2MT \cite{hu2024beyond} utilize specialized attention blocks to aggregate many-to-many viewpoint priors. 
Despite their representational power, the quadratic computational complexity of self-attention leads to excessive computational and memory overhead.
Most recently, State Space Models (SSM), particularly Mamba-based architectures, have emerged to provide global interaction with linear complexity. 

Recent State Space Models (SSM), such as $L^2$FMamba \cite{wei2025l2fmamba} and LFTransMamba \cite{jin2025LFTransMamba}, use selective scanning on serialized SAI and MacPI tokens, while MLFSR \cite{Gao_2024_ACCV} ensures consistency through bi-directional subspace scanning. However, since these spatial-angular interaction methods rely on implicit feature-level correlations rather than explicit epipolar geometry constraints, they often struggle to maintain strict geometric consistency. 
Crucially, these approaches typically employ symmetric quad-directional scanning across all domains, which overlooks the structural anisotropy inherent in different LF representations. 
The lack of domain-specific physical priors in a uniform scanning technique results in considerable computational redundancy and diminishes the effectiveness of high-dimensional geometric modeling.

\subsection{Methods based on Geometric Consistency}
Enforcing explicit geometric constraints is essential for maintaining the structural integrity of LFSR. 
EPIT \cite{Liang_2023_ICCV} tokenizes the EPI into several EPI-stripes to characterize the non-local properties of epipolar geometry. By applying self-attention mechanisms within these stripes, the model effectively models the parallax continuity of long-range geometric structures.

Furthermore, to improve the geometric consistency of spatial-angular interactions, incorporating EPI geometric constraints via multi-representation learning has become a common approach. 
Based on the organizational structure of these representations, existing methods can be classified into parallel, sequential, and cascade paradigms. 

The parallel paradigm, pioneered by the disentangled learning framework in DistgSSR \cite{wang2022disentangling}, utilizes separate branches for spatial, angular, and epipolar features to independently extract domain-specific information before fusion. 
LFMamba \cite{xia2024} further advanced this architecture by incorporating Mamba into these parallel branches to capture long-range dependencies with linear complexity. 
However, such parallel processing often leads to feature redundancy and resource waste because features are extracted independently across domains.

The sequential paradigm prioritizes spatial-angular interaction over the integration of epipolar geometric constraints \cite{wang2024triformer, ke2025CIJSANet, yu2025lfmix}. 
Building on this, HI-LLF \cite{LI2025113240} introduces interleaved interaction units to establish a hierarchical texture base before applying an EPI-specific refinement module. 
However, these methods often accumulate errors during the initial spatial-angular stages, leading to structural misalignments that are difficult to rectify in subsequent geometric optimization steps.

The cascade paradigm adopts a more fine-grained alternating extraction strategy. Zhang et al. \cite{ZHANG2025111616} proposed an adaptive feature aggregation (AFA) framework based on cascade residual learning. This approach sequentially cascades inter-intra spatial (II-SFE), inter-intra angular (II-AFE), and horizontal-vertical epipolar feature extractors within each fundamental aggregation block, repeating this process across $M$ aggregation groups. 
However, the II-SFE and II-AFE modules exhibited high similarity and overlap in feature processing, leading to significant feature redundancy and wasted computation.

Conversely, our proposed RASLF ensures that no redundant information exists between successive refinement steps. Unlike conventional methods that analyze isolated 2D epipolar slices, our approach creates a Panoramic Epipolar Representation to capture global geometric constraints. Consequently, RASLF effectively prevents feature redundancy common in cascaded frameworks and addresses error accumulation common in sequential processing, thereby ensuring strong geometric consistency.

\begin{figure*}[!t]
    \centering
    \includegraphics[width=18cm]{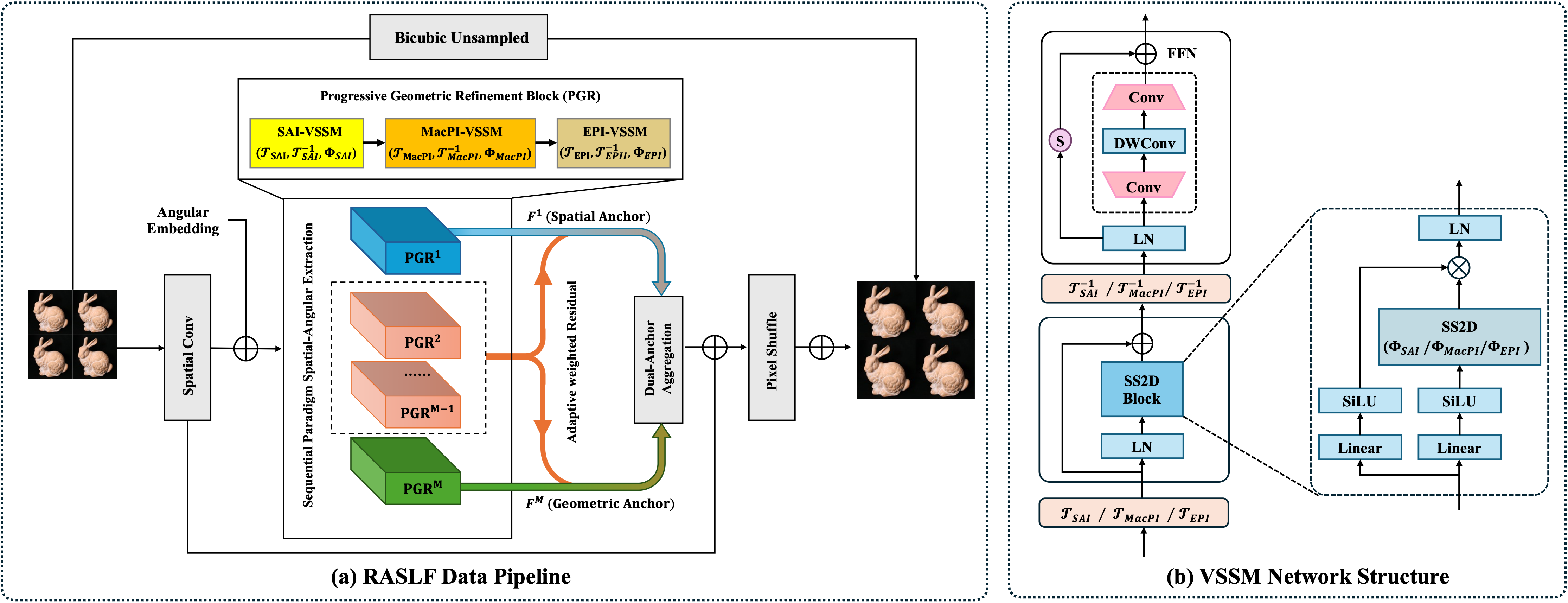}
    \caption{Overall architecture of the proposed RASLF. 
    (a) The Progressive Geometric Refinement (PGR) paradigm sequentially refines features using representation-specific VSSM units, and a Dual-Anchor Aggregation module fuses multi-stage features via spatial and geometric anchors.
    (b) Representation-Aware Asymmetric Scanning (RAAS) tailors SS2D scanning paths $\Phi$ and representation transforms 
    $\mathcal{T}, \mathcal{T}^{-1}$ for SAI, MacPI, and EPI, reducing redundant computation while preserving geometry-aware dependencies.}
    \label{fig:pipeline}
\end{figure*}

\section{Methodology}

\subsection{Overview Architecture}

Let $I_{LR} \in \mathbb{R}^{U \times V \times H \times W}$ denote the input low-resolution (LR) LF image, where $(U, V)$ and $(H, W)$ represent the angular and spatial resolutions, respectively.
The objective of LFSR is to reconstruct a high-resolution (HR) counterpart $I_{HR} \in \mathbb{R}^{U \times V \times \alpha H \times \alpha W}$, where $\alpha$ denotes the upsampling factor.
To effectively restore high-frequency textures while preserving global structural integrity, a residual learning paradigm is adopted.
As illustrated in Fig. 1, a global skip connection is established via a bicubic upsampling operator $\mathcal{B}(\cdot)$, yielding the base LF image $I_{\uparrow} = \mathcal{B}(I_{LR}) \in \mathbb{R}^{U \times V \times \alpha H \times \alpha W}$.
The network is specifically designed to predict the residual component $R$, whereby the final super-resolved light field is formulated as $\hat{I}_{HR} = I_{\uparrow} + R$.

The input $I_{LR}$ is first projected into a latent space via a spatial convolution $\mathcal{S}_{conv}$, followed by a learnable Angular Embedding $P_{ang}$ to encode viewpoint-specific geometric context:
\begin{equation} 
F^0 = \mathcal{S}_{conv}(I_{LR}) + P_{ang}, 
\end{equation}
where $F^0 \in \mathbb{R}^{U \times V \times H \times W \times C}$ serves as the initial state for subsequent hierarchical feature extraction.

The backbone of our architecture consists of $M$ cascaded Progressive Geometric Refinement (PGR) blocks for hierarchical spatio-angular feature extraction.
Each PGR block applies a sequential transformation across the spatial (SAI), angular (MacPI), and epipolar (EPI) representation domains:
\begin{equation} 
F^k = \text{PGR}^k(F^{k-1}), \quad k = 1, \dots, M. 
\end{equation}
This design facilitates step-by-step refinement of spatial textures and angular parallax, leading to a robust, geometrically consistent representation.

Furthermore, a Dual-Anchor Aggregation (DAA) module is developed to generate an aggregated feature representation $F^{agg}$.
By designating the initial and final cascaded features as spatial and geometric anchors, and adding intermediate layers to enhance residuals, the DAA module reduces hierarchical redundancy while maintaining both spatial accuracy and angular consistency.
The total aggregated representation $F^*$ is then generated via a fusion layer:
\begin{equation} 
F^* = F^{agg} + F^0.
\end{equation}
Finally, $F^*$ is processed by a pixel-shuffle upsampling module $\mathcal{P}_{sr}$ to generate the residual $R$, thereby completing the LF reconstruction.
During the training phase, the network is supervised using an L1 loss, 
\begin{equation}
\mathcal{L} = \|\hat{I}_{HR} - I_{HR}\|_1. 
\end{equation}

\subsection{Progressive Geometric Refinement}
The proposed Progressive Geometric Refinement (PGR) block serves as the basic computational unit for transforming LF representations from low-level spatial textures into high-level geometric constraints.
Unlike the common parallel unordered interaction strategies in previous LFSR methods \cite{wang2022disentangling, xia2024}, or the sequential architectures that completely decouple spatial-angular interaction from geometric alignment \cite{wang2024triformer, ke2025CIJSANet, yu2025lfmix}, our method designs a cascade processing chain that explicitly exploits the multi-dimensional physical properties of LF data.

The main motivation is that in traditional sequential architectures, early spatial-angular feature extraction lacks explicit parallax constraints, allowing inconsistent feature responses to spread and accumulate through deep layers.
Conversely, the proposed cascaded strategy interleaves refinement across the SAI, MacPI, and EPI domains within each PGR block, enabling "coupling while calibration."
Performing real-time geometric calibration at each depth level establishes accurate search benchmarks for feature extraction, effectively reducing visual artifacts caused by matching ambiguities and preventing geometric offsets from accumulating in deeper layers.

To ensure algorithmic consistency and computational efficiency, we adopt the visual state space model (VSSM) proposed in \cite{wei2025l2fmamba} as the base operator, whose effectiveness and efficiency have been well proven on the LFSR task.
VSSM enables effective long-range dependency modeling with linear computational complexity, which is essential for processing high-dimensional and redundant LF data.
We define the spatial-angular feature extraction process within the PGR block from $F^{k-1}$ to $F^k$ as a generic processor operator $\mathcal{P}(F, \mathcal{T}, \mathcal{T}^{-1}, \Phi)$.
Here, $F \in \mathbb{R}^{U \times V \times H \times W \times C}$ denotes the input 4D LF feature tensor, $\mathcal{T}(\cdot)$ and $\mathcal{T}^{-1}(\cdot)$ represent the domain-specific transformation and its inverse, respectively. $\Phi$ denotes the predefined set of scanning paths.

Within the specific feature extraction pipeline, the input feature $F^{k-1}$ first undergoes intra-view spatial refinement.
The transformation $\mathcal{T}_{sai}$ reshapes the 4D tensor into tiled spatial slices of dimension $(U \cdot V) \times (H \cdot W) \times C$, which are then processed using a domain-specific scanning set $\Phi_{SAI}$. The resulting intermediate spatial-refined feature is expressed as follows:
\begin{equation}
F_{SAI}^{k-1} = \mathcal{P}(F^{k-1}, \mathcal{T}_{SAI}, \mathcal{T}_{SAI}^{-1}, \Phi_{SAI}).
\end{equation}
Subsequently, the feature is reorganized into the MacPI domain via $\mathcal{T}_{MAC}$, mapping it to a $(H \cdot W) \times (U \cdot V) \times C$ tensor to facilitate angular coupling:
\begin{equation}
F_{MAC}^{k-1} = \mathcal{P}(F_{SAI}^{k-1}, \mathcal{T}_{MAC}, \mathcal{T}_{MAC}^{-1}, \Phi_{MAC}).
\end{equation}
Through this hierarchical preprocessing, the features are endowed with a robust textural foundation and initial parallax awareness before the geometric alignment stage.

\begin{figure}[!t]
    \centering
    \includegraphics[width=\columnwidth]{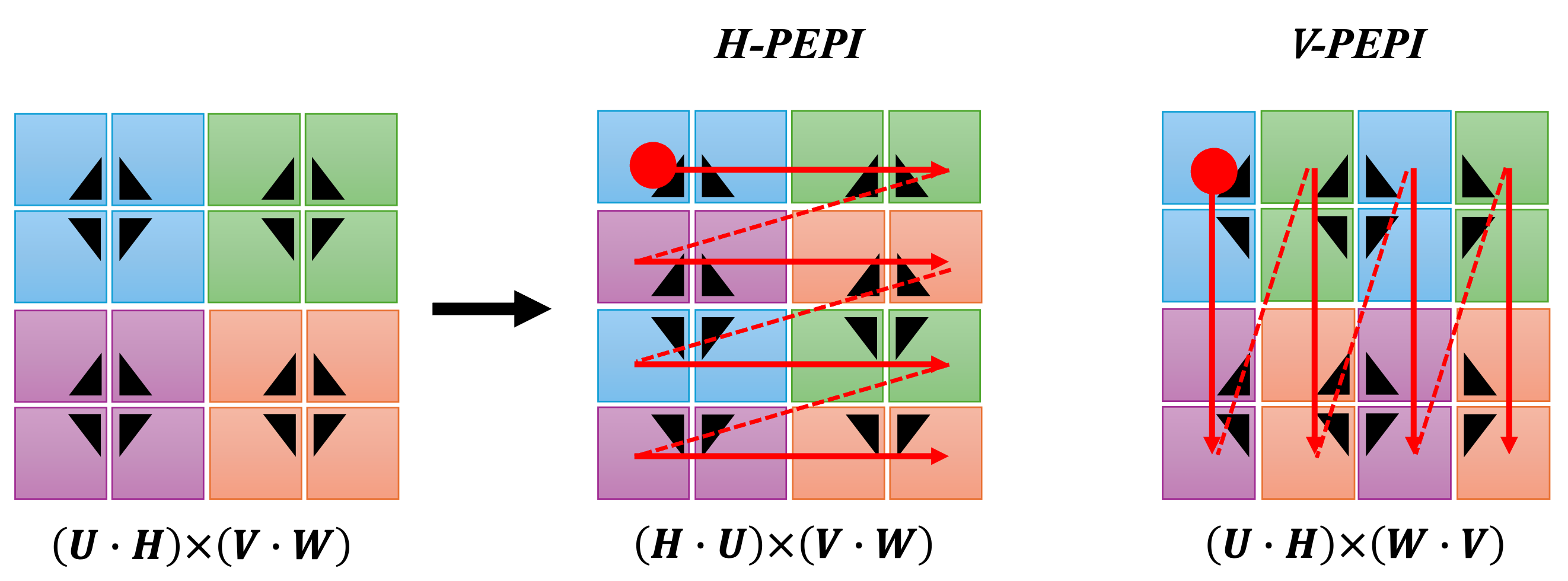}
    \caption{Illustration of the Panoramic Epipolar Representation and the corresponding representation-aware scanning paths (indicated by red solid arrows). }
    \label{fig:PEPI}
\end{figure}

As the key part of the PGR block, the epipolar geometric alignment explicitly enforces the intrinsic linear geometric consistency of the light field. 
Unlike conventional methods that analyze isolated 2D epipolar slices, our approach develops a Panoramic Epipolar (PEPI) representation to capture global geometric constraints.
The composite epipolar transformation $\mathcal{T}_{epi}$ comprises two symmetric reorganization branches: the Vertical Panoramic EPI (V-PEPI), generated via $\mathcal{T}_{v}$ as a $(H \cdot U) \times (V \cdot W) \times C$ tensor, and the Horizontal Panoramic EPI (H-PEPI), concurrently reshaped via $\mathcal{T}_{h}$ into a $(W \cdot V) \times (U \cdot H) \times C$ representation (see Fig. \ref{fig:PEPI}).

This innovative tiling approach explicitly and effectively maps parallax-related information, originally dispersed in the 4D spatio-angular domain, onto structured 2D planes. 
This globalized representation allows the processor to not only capture linear slopes within individual epipolar lines but also to observe structural correlations across different spatial locations, utilizing the long-range modeling ability of SSMs.
Ultimately, the geometric features are integrated and processed, with the output of the PGR block, $F^k$, expressed as follows:
\begin{equation}
\begin{aligned}
    F^k = \mathcal{T}_{epi}^{-1}(\mathcal{P}( & \hat{F}_{mac}, \mathcal{T}_{epi}, \mathcal{T}_{epi}^{-1}, \Phi_{epi})), \\
    \text{where} \quad \mathcal{T}_{epi} = \{ & \mathcal{T}_{h}, \mathcal{T}_{v}\} \quad \text{and} \quad \Phi_{epi} = \{\Phi_{h}, \Phi_{v}\}.
\end{aligned}
\end{equation}

Compared to the implicit modeling in \cite{wei2025l2fmamba}, which scans the angular grid sequentially, our scheme mitigates information decay during long-range state updates and, crucially, circumvents the geometric constraint lag inherent in the "coupling before alignment" sequence of traditional architectures.
By jointly characterizing spatial positions and view variations on a unified EPI plane within each block, parallax structures are calibrated in real-time in a more concentrated and explicit manner.

\subsection{Representation-Aware Asymmetric Scanning Strategy}
\label{section: RAAS}

In two-dimensional (2D) image state-space modeling, scanning directions are used to serialize spatial pixels into sequential state-update paths, capturing directional dependencies.
Current methods \cite{liu2024vmamba, guo2024mambair} typically adopt the 2D Selective Scanning (SS2D) mechanism, which employs a quad-directional scanning configuration $\Phi_{4\text{-}path} = \{\Phi_{row}, \Phi_{row}^{'}, \Phi_{col}, \Phi_{col}^{'}, \}$ across all feature domains, representing forward/inverse row-major and column-major trajectories.
However, we argue that applying this quad-directional scanning uniformly across all LF representations ignores their distinct physical priors and introduces unnecessary computational redundancy. 
To address this issue, we proposed the Representation-Aware Asymmetric Scanning (RAAS) strategy, which dynamically tailors the scanning path set $\Phi$ to the physical characteristics of disparate representation domains, achieving an optimal balance between reconstruction performance and computational efficiency through strategic path pruning.

In the intra-view spatial refinement stage, SAI-VSSM aims to capture local spatial dependencies within individual sub-aperture images.
Compared with the strong directional structures in MacPI and EPI representations, SAIs exhibit more locally symmetric spatial dependencies, making simple forward scanning sufficient for effective neighborhood correlation modeling.
In this case, backward scanning tends to capture largely overlapping contextual dependencies, leading to limited additional benefits while introducing extra computational overhead.
Consequently, we implemented a path pruning strategy in SAI-VSSM by retaining only forward state-update paths, i.e., $\Phi_{sai} = \{\Phi_{row}, \Phi_{col}\}$. 

Unlike SAI, MacPI interleaves spatial and angular dimensions, such that neighboring elements along each scan axis no longer represent simple local adjacency, but instead reflect coupled relations across different views and spatial positions. As a result, forward and backward scanning propagate context through different dependency chains and capture distinct yet complementary spatial-angular correlations, making bidirectional scanning necessary rather than redundant.
Thus, MacPI-VSSM preserves the full quad-directional scanning set, i.e., $\Phi_{mac} = \Phi_{4\text{-}path}$, to ensure adequate inter-view correspondence coupling.

The path design for EPI-VSSM is motivated by the strong directional structure of epipolar representations. In our PEPI representation, epipolar trajectories are consistently aligned with the row axis in H-PEPI and the column axis in V-PEPI. Under this formulation, the SSM can effectively propagate parallax-dependent information along these physically meaningful linear paths, making a single forward scan sufficient to capture the dominant geometric dependencies. In contrast, scanning along other directions provides limited additional geometric cues, as they do not follow the principal orientation of epipolar trajectories and mainly introduce redundant context modeling. 
Therefore, for the decoupled horizontal and vertical branches, we reduce the scanning paths to a single direction $\Phi_{v\text{-}epi} = \{\Phi_{col}\}$ and $\Phi_{h\text{-}epi} = \{\Phi_{row}\}$, as shown in Fig. \ref{fig:PEPI}.

From a computational complexity perspective, the RAAS strategy compresses the structure of the state-space modeling branches by matching the scanning density to the needs of physical modeling.
Since each scanning path represents an independent state-update operation, reducing the number of directions directly lowers the parameter count and computational cost. 
As a result, our method attains greater computational efficiency and stronger geometric consistency representations while still preserving adequate modeling capacity. 

\subsection{Dual-Anchor Aggregation Module}
In cascaded LFSR architectures, the deliberate integration of hierarchical features is crucial for enhancing reconstruction accuracy.
Typically, the evolution of features within these cascaded structures shows a clear functional hierarchy.
The shallow features tend to retain the local spatial textures of the original input, whereas deep features, through receptive-field expansion and nonlinear mapping, gradually develop into global representations that incorporate spatial-angular geometric information.
To fully utilize these complementary characteristics without creating hierarchical redundancy, we propose the Dual-Anchor Aggregation (DAA) mechanism.

LFSR requires balancing the accuracy of high-frequency details with the stability of angular structures; to achieve this, the DAA module explicitly sets the endpoints of the cascading path as main reference points.
Since the initial features $F^1$ have not undergone extensive abstraction and retain the most intact original spatial details, they are designated as the anchor for spatial texture restoration. Meanwhile, the final features $F^M$, which integrate spatial-angular geometric constraints, are regarded as the anchor for global geometric modeling.

To enable effective utilization of intermediate feature information within the cascaded sequence $\{F^k\}_{k=2}^{M-1}$, we treat them as adaptive refinement operators. 
By injecting progressive information from intermediate layers into the boundary anchors through weighted residuals, we constructed enhanced spatial $F^S$ and geometric $F^G$ anchors:
\begin{equation} 
F^S = w^S_1 \cdot F^1 + \sum_{k=2}^{M-1} w^S_k \cdot F^k, \end{equation}
\begin{equation} 
F^G = w^G_M \cdot F^M + \sum_{k=2}^{M-1} w^G_k \cdot F^k, \end{equation}
Where $\{w^S_k\}_{k=1}^{M-1}$ and $\{w^G_k\}_{k=2}^{M}$ denote the decoupled weighting coefficients that regulate the contribution of each layer to the respective anchors.

This design enables thorough recovery of feature resources and maximizes effectiveness, thereby improving the texture expressiveness of the spatial anchor and strengthening the structural robustness of the geometric anchor.
Subsequently, the DAA module uses dimensional concatenation and a projection layer to achieve deep coupling of these two complementary anchor representations, generating the final aggregated features $F_{agg}$:
\begin{equation} 
F^{agg} = \text{MLP}(\text{Concat}(F^S, F^G)). 
\end{equation}

This anchor-guided aggregation method explicitly removes the hierarchical redundancy caused by cascading schemes at the structural level.
Instead of simply concatenating all layers, our approach uses intermediate features as directional refinements, keeping the reconstruction rooted in stable spatial references and accurate geometric benchmarks.

\begin{table*}[!t]
    \centering
    \caption{PSNR/SSIM results compared with SOTA methods for 2$\times$ and 4$\times$ LFSR tasks. 
             The best and second-best results are, respectively, in bold and underlined.}
    \begin{tabular}{@{}cccccccc@{}}
        \toprule[2pt]
        Method                                  & Scale        & EPFL           & HCINew         & HCIold         & INRIA          & STFgantry      & Average        \\ 
        \midrule[1pt]  
        RCAN        \cite{zhang2018image}        & $\times$2    & 33.156/0.9635  & 35.022/0.9603  & 41.125/0.9875  & 35.036/0.9769  & 36.670/0.9831  & 36.202/0.9743  \\
        resLF       \cite{zhang2019residual}     & $\times$2    & 33.617/0.9706  & 36.685/0.9739  & 43.422/0.9932  & 35.395/0.9804  & 38.354/0.9904  & 37.495/0.9817  \\
        LFSSR       \cite{yeung2018light}        & $\times$2    & 33.671/0.9744  & 36.802/0.9749  & 43.811/0.9938  & 35.279/0.9832  & 37.944/0.9898  & 37.501/0.9832  \\ 
        LF-ATO      \cite{jin2020light}          & $\times$2    & 34.272/0.9757  & 37.244/0.9767  & 44.205/0.9942  & 36.171/0.9842  & 39.636/0.9929  & 38.306/0.9847  \\
        MEG-Net     \cite{zhang2021end}          & $\times$2    & 34.312/0.9773  & 37.424/0.9777  & 44.097/0.9942  & 36.103/0.9849  & 38.767/0.9915  & 38.141/0.9851  \\
        DistgSSR    \cite{wang2022disentangling} & $\times$2    & 34.809/0.9787  & 37.959/0.9796  & 44.943/0.9949  & 36.586/0.9859  & 40.404/0.9942  & 38.940/0.9867  \\
        LF-InterNet \cite{wang2020spatial}       & $\times$2    & 34.112/0.9760  & 37.170/0.9763  & 44.573/0.9946  & 35.829/0.9843  & 38.435/0.9909  & 38.024/0.9844  \\
        LF-IINet    \cite{liu2021intra}          & $\times$2    & 34.736/0.9773  & 37.768/0.9790  & 44.852/0.9948  & 36.564/0.9853  & 39.894/0.9936  & 38.763/0.9860  \\
        HLFSR-SSR   \cite{vinh2023-lfsr}         & $\times$2    & 35.310/0.9800  & 38.317/0.9807  & 44.978/\underline{0.9950}  & 37.060/\underline{0.9867}  & 40.849/0.9947  & 39.303/0.9874  \\
        DPT         \cite{wang2022detail}        & $\times$2    & 34.490/0.9758  & 37.355/0.9771  & 44.302/0.9943  & 36.409/0.9843  & 39.429/0.9926  & 38.397/0.9848  \\
        LFT         \cite{liang2022light}        & $\times$2    & 34.783/0.9776  & 37.766/0.9788  & 44.628/0.9947  & 36.539/0.9853  & 40.408/0.9941  & 38.825/0.9861  \\
        EPIT        \cite{Liang_2023_ICCV}       & $\times$2    & 34.826/0.9775  & 38.228/\underline{0.9810}  & \underline{45.075}/0.9949  & 36.672/0.9853  & \textbf{42.166}/0.9957  & 39.393/0.9869  \\
        LF-DET      \cite{cong2023lfdet}         & $\times$2    & 35.262/0.9797  & 38.314/0.9807  & 44.986/\underline{0.9950}  & 36.949/0.9864  & 41.762/0.9855  & 39.455/0.9874  \\
        MLFSR  \cite{Gao_2024_ACCV}       & $\times$2    & 35.218/\underline{0.9801}  & 38.140/0.9803  & 44.904/\underline{0.9950}  & 36.919/0.9865  & 40.975/0.9949 & 39.231/0.9873 \\
        LFMamba  \cite{xia2024}    & $\times$2    & \textbf{35.758}/\textbf{0.9824}  & \underline{38.368}/0.9801  & 44.985/\underline{0.9950}  & \underline{37.063}/\textbf{0.9876}  & 40.954/0.9948 & 39.424/\textbf{0.9881} \\
        $L^2$FMamba \cite{wei2025l2fmamba} & $\times$2    & \underline{35.515}/0.9796  & 38.225/0.9803  & 44.953/0.9949  & \textbf{37.165}/0.9862  & 41.567/\underline{0.9952}  & \underline{39.485}/0.9873  \\
        RASLF (Our)  & $\times$2    & 35.176/0.9791  & \textbf{38.427}/\textbf{0.9813}  & \textbf{45.312}/\textbf{0.9952}  & 36.987/0.9861  & \underline{41.873}/\textbf{0.9955}  & \textbf{39.555}/\underline{0.9875}  \\
        
        \midrule[1pt]  
        \midrule[1pt]  
        
        RCAN        \cite{zhang2018image}        & $\times$4    & 27.904/0.8863  & 29.694/0.8886  & 35.359/0.9548  & 29.800/0.9276  & 29.021/0.9131  & 30.355/0.9141  \\
        resLF       \cite{zhang2019residual}     & $\times$4    & 28.260/0.9035  & 30.723/0.9107  & 36.705/0.9682  & 30.338/0.9412  & 30.191/0.9372  & 31.243/0.9322  \\
        LFSSR       \cite{yeung2018light}        & $\times$4    & 28.596/0.9118  & 30.928/0.9145  & 36.907/0.9696  & 30.585/0.9467  & 30.570/0.9426  & 31.517/0.9370  \\ 
        LF-ATO      \cite{jin2020light}          & $\times$4    & 28.514/0.9115  & 30.880/0.9135  & 36.999/0.9699  & 30.710/0.9484  & 30.607/0.9430  & 31.542/0.9373  \\
        MEG-Net     \cite{zhang2021end}          & $\times$4    & 28.749/0.9160  & 31.103/0.9177  & 37.287/0.9716  & 30.674/0.9490  & 30.771/0.9453  & 31.717/0.9399  \\
        DistgSSR    \cite{wang2022disentangling} & $\times$4    & 28.992/0.9195  & 31.380/0.9217  & 37.563/0.9732  & 30.994/0.9519  & 31.649/0.9534  & 32.116/0.9439  \\
        LF-InterNet \cite{wang2020spatial}       & $\times$4    & 28.812/0.9162  & 30.961/0.9161  & 37.150/0.9716  & 30.777/0.9491  & 30.365/0.9409  & 31.613/0.9388  \\
        LF-IINet    \cite{liu2021intra}          & $\times$4    & 29.048/0.9188  & 31.331/0.9208  & 37.620/0.9734  & 31.039/0.9515  & 31.261/0.9502  & 32.060/0.9429  \\
        HLFSR-SSR   \cite{vinh2023-lfsr}         & $\times$4    & 29.196/0.9222  & 31.571/0.9238  & 37.776/0.9742  & 31.241/0.9543  & 31.641/0.9537  & 32.285/0.9456  \\
        DPT         \cite{wang2022detail}        & $\times$4    & 28.939/0.9170  & 31.196/0.9188  & 37.412/0.9721  & 30.964/0.9503  & 31.150/0.9488  & 31.932/0.9414  \\
        LFT         \cite{liang2022light}        & $\times$4    & 29.261/0.9209  & 31.433/0.9215  & 37.633/0.9735  & 31.218/0.9524  & 31.794/0.9543  & 32.268/0.9445  \\
        EPIT        \cite{Liang_2023_ICCV}       & $\times$4    & 29.339/0.9197  & 31.511/0.9231  & 37.677/0.9737  & 31.372/0.9526  & 32.179/0.9571  & 32.416/0.9452  \\
        LF-DET      \cite{cong2023lfdet}         & $\times$4    & 29.473/0.9230  & 31.558/0.9235  & 37.843/0.9744  & 31.388/0.9534  & 32.139/0.9573  & 32.480/0.9463  \\
        MLFSR  \cite{Gao_2024_ACCV}              & $\times$4    & 29.283/0.9218  & 31.564/0.9235  & 37.831/0.9745  & 31.241/0.9531  & 32.031/0.9567 & 32.389/0.9235 \\
        LFMamba  \cite{xia2024}           & $\times$4    & \textbf{29.840}/\textbf{0.9256}  & \textbf{31.695}/\textbf{0.9249}  & \textbf{37.912}/\textbf{0.9748}  & \textbf{31.808}/\textbf{0.9551}  & 31.846/0.9553 & 32.620/\underline{0.9471} \\
        $L^2$FMamba \cite{wei2025l2fmamba}        & $\times$4    & 29.681/0.9233  & 31.647/0.9243  & 37.864/0.9745  & 31.728/0.9543  &  \underline{32.198}/ \underline{0.9574} & \underline{32.623}/0.9468  \\
        RASLF (Our)  & $\times$4    & \underline{29.763}/\underline{0.9239}  & \underline{31.667}/\underline{0.9245}  & \underline{37.894}/\underline{0.9745}  & \underline{31.758}/\underline{0.9544}  & \textbf{32.368}/\textbf{0.9586}  &  \textbf{32.690}/ \textbf{0.9472} \\
        \bottomrule[2pt]
    \end{tabular}
    \label{tab:quantitative}
\end{table*}

\section{Experiments}

\subsection{Datasets and Implementation Details}
To comprehensively evaluate the effectiveness of RASLF, we conducted experiments on three real-world LF datasets, EPFL \cite{rerabek2016new}, INRIA \cite{le2018light}, and STF-gantry \cite{vaish2008new}, as well as two synthetic LF datasets, HCIold \cite{wanner2013datasets}, and HCInew \cite{honauer2017dataset}. 
For all datasets, we strictly followed the publicly available data splitting protocols and pre-processing procedures adopted in prior work to ensure fair and reproducible comparisons. 
Specifically, we used the central 5$\times$5 views of each LF image and cropped HR patches from these views for both training and testing.
HR patches are cropped with sizes of 64$\times$64 and 128$\times$128 for the 2$\times$ and 4$\times$ LFSSR tasks, respectively.
The corresponding LR inputs are generated via bicubic down-sampling, yielding LR patches of size 32$\times$32.
For quantitative evaluation, we converted the LF images from RGB to YCbCr and computed the peak signal-to-noise ratio (PSNR) and structural similarity (SSIM) on the Y channel only. 
For each dataset, we first averaged the metrics across all LF scenes, then averaged the results across the five datasets to obtain the final scores.

To facilitate the RAAS strategy while avoiding sub-optimal convergence inherent in training sparse architectures from scratch, a two-stage training paradigm is adopted.
First, a full-path model is pre-trained for 180 epochs to capture dense feature representations.
We used the Adam optimizer and a StepLR scheduler, with an initial learning rate of 2e-4 that is decreased by a factor of 0.5 every 30 epochs. 
The training data is augmented with random horizontal and vertical flips and 90-degree rotations. 
Subsequently, guided by the RAAS strategy, redundant scanning paths are pruned according to the representation-specific requirements, as defined in Sec. \ref{section: RAAS}. The pruned model is then fine-tuned for 30 epochs with a learning rate of 5e-5, which is decayed by a factor of 0.5 every 15 epochs to smoothly recover and optimize the final performance.

\subsection{Comparisons with State-of-The-Art Methods}

To thoroughly assess the proposed RASLF, we compared it against 16 representative state-of-the-art methods across quantitative metrics, visual quality, and computational efficiency.
The compared methods cover both SISR and LFSR tasks. 
Specifically, RCAN \cite{zhang2018image} serves as the CNN-based SISR method. 
For LFSR, DPT \cite{wang2022detail}, \cite{liang2022light}, EPIT \cite{Liang_2023_ICCV}, and LF-DET \cite{cong2023lfdet} are built upon Transformer architectures, whereas LFMamba \cite{xia2024}, $L^2$FMamba \cite{wei2025l2fmamba}, and MLFSR \cite{Gao_2024_ACCV} adopt state-space structures. 
The remaining approaches are based on CNN architectures, including LFSSR \cite{yeung2018light}, MEG-Net \cite{zhang2021end}, LF-ATO \cite{jin2020light}, LF-IINet \cite{liu2021intra}, DistgSSR \cite{wang2022disentangling}, and HLFSR-SSR \cite{vinh2023-lfsr}.

\begin{figure*}[!t]
    \centering
    \includegraphics[width=18cm]{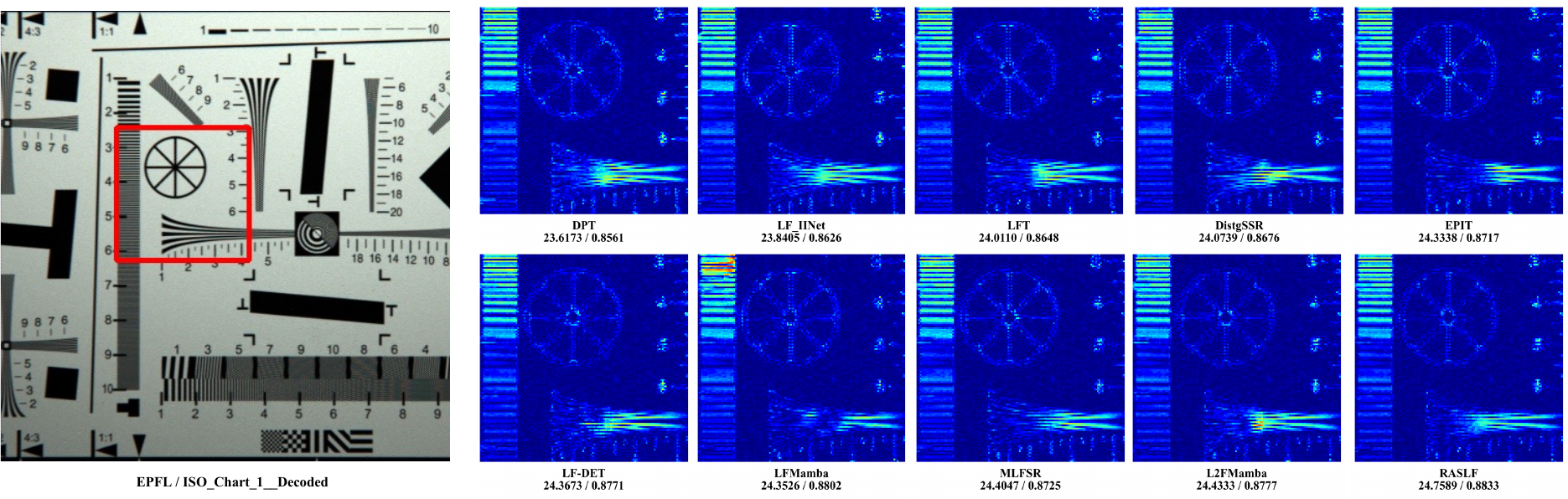}
    \quad
    \includegraphics[width=18cm]{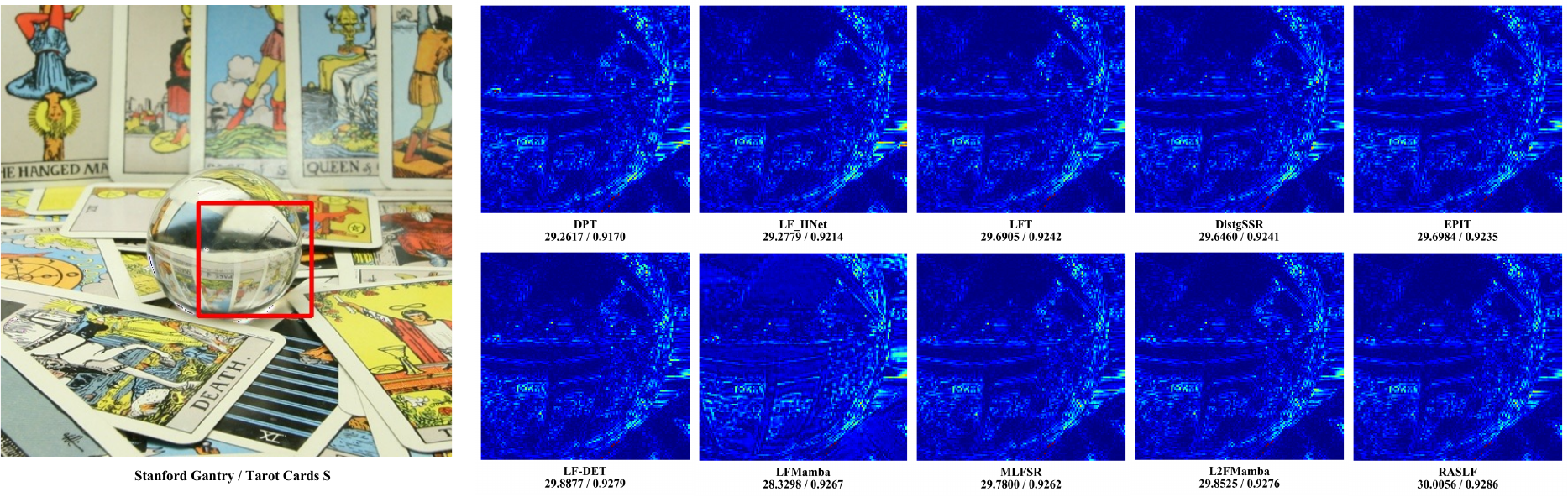}
    \caption{
            Qualitative visualization results for 4$\times$ LFSR compared to other methods. 
            Here, we showed the error maps of the reconstructed center-view images, with representative regions indicated by arrows.
            PSNR/SSIM values for the corresponding region are provided below.}
    \label{fig:4x}
\end{figure*}

\begin{table*}[!t]
    \fontsize{6.5pt}{8pt}\selectfont
    \centering
    \caption{Comparison of parameters, flops, time, and average PSNR/SSIM values for $\times$2 and $\times$4 SR. Flops and time are calculated on an input LF with a size of $5\times5\times32\times32$.}
    
    \begin{minipage}[t]{0.45\textwidth}
    \centering
    \begin{tabular}
    {@{}c@{\hspace{3pt}}c@{\hspace{4pt}}c@{\hspace{3pt}}c@{\hspace{3pt}}c@{\hspace{3pt}}c@{}}
        \toprule[2pt]
        Method                                  & Scale         & Params.     & FLOPs(G)     & Time(ms)       & {\color{black}{Avg. PSNR/SSIM}}  \\ 
        \midrule[1pt]  
        LFSSR      \cite{yeung2018light}        & $\times$2     & 0.89M       & 25.70        & 10.0           & 37.501/0.9832   \\
        MEG-Net    \cite{zhang2021end}          & $\times$2     & 1.69M       & 48.40        & 31.2           & 38.141/0.9851   \\
        LF-ATO     \cite{jin2020light}          & $\times$2     & 1.22M       & 597.66       & 85.6           & 38.306/0.9847   \\
        LF-IINet   \cite{liu2021intra}          & $\times$2     & 5.04M       & 56.16        & 20.6           & 38.763/0.9860   \\
        DistgSSR   \cite{wang2022disentangling} & $\times$2     & 3.53M       & 64.11        & 24.2           & 38.940/0.9867   \\
        HLFSR-SSR  \cite{vinh2023-lfsr}         & $\times$2     & 13.72M      & 167.81       & 31.1           & 39.303/0.9874   \\
        LFT        \cite{liang2022light}        & $\times$2     & 1.11M       & 56.16        & 91.4           & 38.825/0.9861   \\
        DPT        \cite{wang2022detail}        & $\times$2     & 3.73M       & 65.34        & 98.5           & 38.397/0.9848   \\
        EPIT       \cite{Liang_2023_ICCV}       & $\times$2     & 1.42M       & 69.71        & 32.2           & 39.393/0.9869   \\
        LF-DET     \cite{cong2023lfdet}         & $\times$2     & 1.59M       & 48.50        & 65.9           & 39.455/0.9874   \\
        {\color{black}{MLFSR  \cite{Gao_2024_ACCV}}}       & {\color{black}{$\times$2}}    & {\color{black}{1.36M}}  & {\color{black}{53.30}}  & {\color{black}{27.8}}  & {\color{black}{39.231/0.9873}} \\
        {\color{black}{LFMamba \cite{xia2024}}}     & {\color{black}{$\times$2}}    & {\color{black}{2.15M}}  & {\color{black}{92.29}}  & {\color{black}{75.9}}  & {\color{black}{39.424/0.9881}} \\
        $L^2$FMamba \cite{wei2025l2fmamba}                    & $\times$2     & 1.04M       & 36.59        & 31.9           & 39.485/0.9873   \\
         RASLF (Our)                      & $\times$2     & 0.90M       & 29.77        & 28.8           & 39.552/0.9876   \\
        \bottomrule[1pt]
        \end{tabular}
    \end{minipage}%
    \begin{minipage}[t]{0.45\textwidth}
    \centering
    \begin{tabular}
{@{}c@{\hspace{3pt}}c@{\hspace{4pt}}c@{\hspace{3pt}}c@{\hspace{3pt}}c@{\hspace{3pt}}c@{}}
        \toprule[2pt]
        Method                                  & Scale         & Params.     & FLOPs(G)     & Time(ms)       & {\color{black}{Avg. PSNR/SSIM}}  \\ 
        \midrule[1pt]  
        LFSSR      \cite{yeung2018light}        & $\times$4     & 1.61M       & 128.44       & 37.7           & 31.517/0.9370   \\
        MEG-Net    \cite{zhang2021end}          & $\times$4     & 1.77M       & 102.20       & 32.4           & 31.717/0.9399   \\
        LF-ATO     \cite{jin2020light}         & $\times$4     & 1.66M       & 686.99       & 88.7           & 31.542/0.9373   \\
        LF-IINet   \cite{liu2021intra}         & $\times$4     & 4.89M       & 57.42        & 20.8           & 32.060/0.9429   \\
        DistgSSR   \cite{wang2022disentangling} & $\times$4     & 3.58M       & 65.41        & 25.0           & 32.116/0.9439   \\
        HLFSR-SSR  \cite{vinh2023-lfsr}         & $\times$4     & 13.87M      & 182.93       & 32.7           & 32.285/0.9456  \\
        LFT        \cite{liang2022light}       & $\times$4     & 1.16M       & 57.60        & 95.2           & 32.268/0.9445   \\
        DPT        \cite{wang2022detail}        & $\times$4     & 3.78M       & 66.55        & 99.7           & 31.932/0.9414   \\
        EPIT       \cite{Liang_2023_ICCV}       & $\times$4     & 1.47M       & 71.15        & 33.6           & 32.416/0.9452   \\
        LF-DET     \cite{cong2023lfdet}         & $\times$4     & 1.69M       & 51.20        & 75.0           & 32.480/0.9463   \\
        {\color{black}{MLFSR  \cite{Gao_2024_ACCV}}}       & {\color{black}{$\times$4}}    & {\color{black}{1.41M}}  & {\color{black}{54.74}}  & {\color{black}{28.9}}  & {\color{black}{32.389/0.9235}} \\
        {\color{black}{LFMamba  \cite{xia2024}}}    & {\color{black}{$\times$4}}    & {\color{black}{2.30M}}  & {\color{black}{96.24}}  & {\color{black}{77.1}}  & {\color{black}{32.620/0.9471}} \\
        $L^2$FMamba    \cite{wei2025l2fmamba}                 & $\times$4     & 1.09M       & 37.99        & 32.6           & 32.623/0.9468   \\
         RASLF (Our)                      & $\times$4     & 0.95M       & 31.20        & 29.4           & 32.690/0.9472   \\
        \bottomrule[1pt]
    \end{tabular}
    \end{minipage}
    
    \label{tab:efficiency}
\end{table*}

\begin{enumerate}[wide]

    \item{
        \emph{Quantitative Results:} 
        The quantitative comparisons between RASLF and other state-of-the-art methods are presented in Table \ref{tab:quantitative}.
In the $4\times$ LFSR task, RASLF achieves the best performance across all datasets, consistently outperforming the efficient state-space baseline $L^2$FMamba.
Notably, on the STF-gantry dataset, which features large-scale parallax, RASLF surpasses $L^2$FMamba by 0.17 dB and LFMamba by 0.52 dB. This significant improvement confirms that our PGR block effectively enforces explicit geometric calibration on panoramic representations, enabling precise modeling of complex cross-view dependencies.
In the $2\times$ LFSR task, RASLF remains highly competitive, achieving the highest average PSNR among all methods. Although LFMamba achieves a slightly higher SSIM, RASLF still outperforms its direct competitor, $L^2$FMamba, in both PSNR and SSIM.
Considering the extremely low parameter count of RASLF, these results indicate that our method achieves a favorable balance between pixel-level fidelity and geometric consistency while maintaining efficient inference.
    }
    
    \item{
        \emph{Qualitative Results:} 
        Fig. \ref{fig:4x} presents the qualitative results of error maps for different methods on the $4\times$ LFSR task. 
        Compared with other state-of-the-art methods, our approach shows better ability to restore texture details and maintain structural consistency. 
This is primarily attributed to our proposed PGR block and Panoramic Epipolar Representation, which jointly reorganize fragmented local epipolar constraints into a globally coherent geometric structure, as evidenced by the significantly lower error responses. In varied and challenging scenes, such as the dense radial lines in "EPFL/ISO\_Chart\_1\_Decoded" and the complex specular reflections and distinct contours of the crystal ball in "Stanford Gantry/Tarot Cards S", our method achieves outstanding performance, producing the cleanest error maps.
In particular, compared to recent SSM-based architectures such as LFMamba and $L^2$FMamba, our proposed RASLF achieves more visually pleasing error suppression and superior quantitative results.
    }
    
    \begin{figure}[!t]
    \centering
    \includegraphics[width=0.9\linewidth]{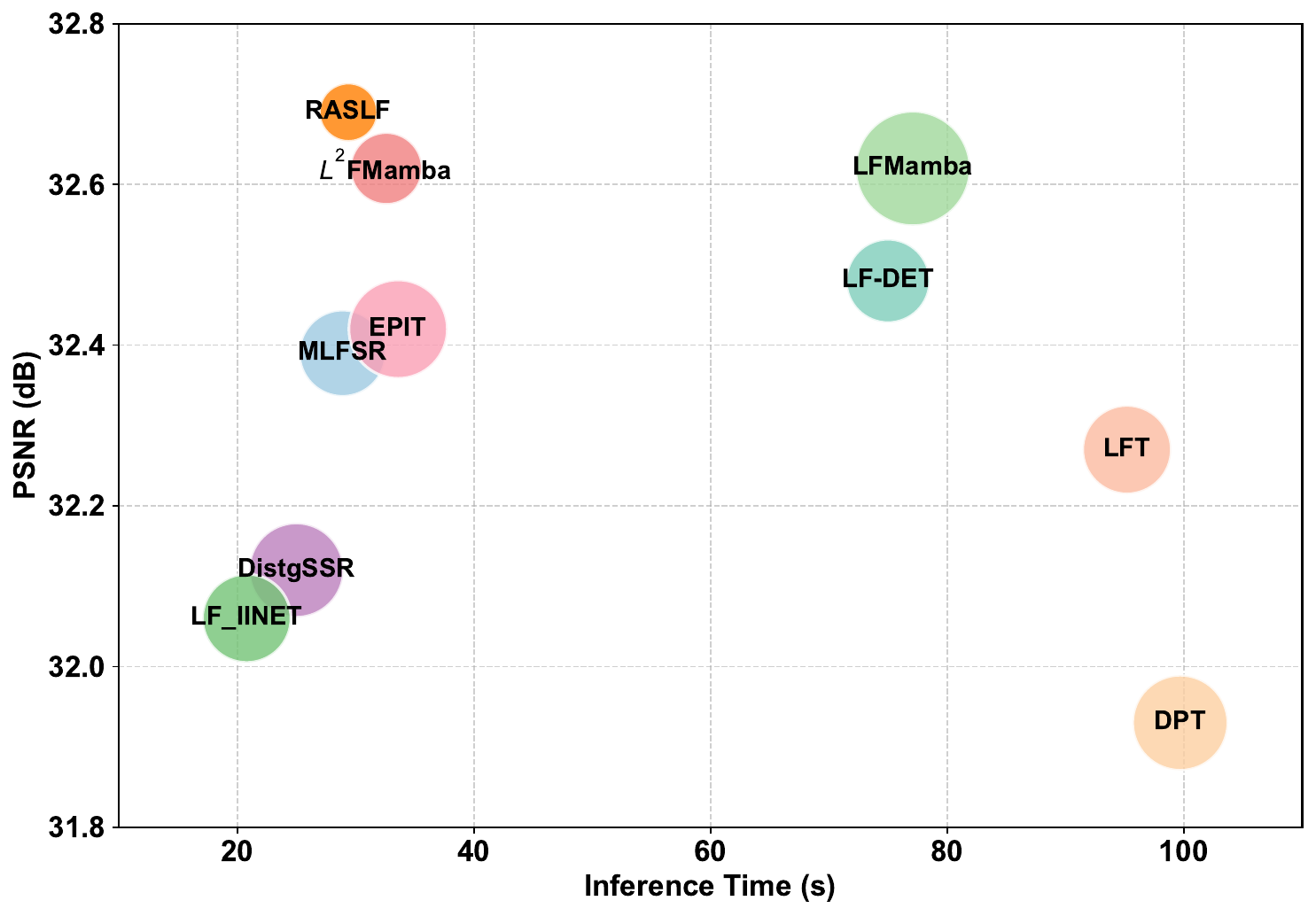}
    \caption{{\color{black}{Comparison of computational efficiency between our method and SOTA models on the $4\times$ LFSR task. The area of each circle represents memory consumption.}}}
    \label{fig:bubble_chart}
    \end{figure}

    \begin{table}[!t]
    \centering
    \caption{Ablation study on different components for 4$\times$ LFSR.}
    \begin{tabular}{cccc}
    \toprule[2pt]
      &  PSNR / SIMM &  Params & FLOPS\\
     \midrule[1pt]  
      w./o. DDA           &32.637 / 0.9466 &   1.086 & 37.928 \\
      \midrule[1pt]
      w. EPI-H \& EPI-V     &32.588 / 0.9466 &   0.953 & 31.200 \\
      w. EPIs-H \& EPIs-V   &32.628 / 0.9468 &   0.953 & 31.200 \\
      \midrule[1pt]
      RASLF (Our)   &32.690 / 0.9472 &   0.953 & 31.200 \\
      
    \toprule[2pt]
    \end{tabular}
    \label{table:components}
    \end{table}
    
    \item{\emph{Computational Efficiency:} 
    We compared RASLF with representative existing methods under standardized conditions in terms of model complexity, including the number of parameters (Params.), floating-point operations (FLOPs), and inference time (Time).
    For the LFSR task, all methods took LF patches of size $5\times5\times32\times32$ as input and are evaluated on a unified NVIDIA RTX 3090 GPU hardware platform, covering both $\times2$ and $\times4$ SR scales.
    Quantitative efficiency metrics in Table \ref{tab:efficiency} reveal that RASLF achieves a superior trade-off between model complexity and accuracy.

Specifically, compared to Transformer-based methods, RASLF significantly lowers computational costs and provides much faster inference. 
Although its inference speed is slightly lower than that of lightweight CNN-based methods, it remains highly competitive regarding overall efficiency.
Notably, current state-space model implementations are not yet as heavily optimized for GPUs as convolutional operators; therefore, with continued development of hardware-aware SSM operators, the inference efficiency of RASLF is expected to improve further.

Compared with MLFSR and LFMamba, RASLF achieves higher reconstruction accuracy with fewer parameters, lower computational complexity, and shorter inference time.
Furthermore, even compared with the state-of-the-art efficiency-oriented model $L^2$FMamba, which adopts a similar VSSM backbone, RASLF maintains a clear advantage.
Specifically, at the $4\times$ scale, RASLF achieves a 12.8\% reduction in parameter scale and a 17.9\% reduction in FLOPs compared to $L^2$FMamba, while also improving the PSNR by 0.07 dB. 
This shows that pruning strategic paths in our RAAS approach effectively removes unnecessary calculations without losing representational capability.

To evaluate runtime efficiency and resource usage, we compared the inference time and GPU memory consumption between our method and SOTA models on the $4\times$ LFSR task in Fig. \ref{fig:bubble_chart}. The area of each circle indicates peak memory usage. Our proposed RASLF achieves a superior balance between reconstruction performance and computational cost. By effectively pruning hierarchical and directional redundancies, our framework maximizes resource utilization for critical reconstruction information.
    }
\end{enumerate}

\subsection{Ablation Study}

\subsubsection{Dual-Anchor Aggregation Module}

To verify the effectiveness of the Dual-Anchor Aggregation (DAA) module in hierarchical feature fusion, we evaluate a variant in which the DAA is replaced with a simple feature concatenation operation, denoted w/o DAA.
As reported in Table \ref{table:components}, removing DDA increases the parameter count by about 14.0\% and raises FLOPs by about 21.6\%, while the reconstruction accuracy decreases by 0.05 dB in PSNR.
This contrast confirms that simple hierarchical stacking introduces significant hierarchical redundancy, whereas the DAA effectively improves feature utilization and reduces model complexity through an anchor-guided mechanism.

\subsubsection{Panoramic Epipolar Representation}
We further investigated the impact of different epipolar representations on reconstruction performance.
While keeping the rest of the network fixed, the Panoramic Epipolar Representation (PER) is compared against conventional isolated epipolar slices (w. EPI) and stacked epipolar slices (w. EPIs).

As shown in Table \ref{table:components}, the configuration with isolated EPI slices delivers the lowest reconstruction accuracy, achieving a PSNR of 32.588 dB.
This is because fragmented slices interrupt the geometric continuity of the light field, preventing the capture and transmission of global disparity dependencies.
While stacking EPIs partially alleviates this issue and improves PSNR to 32.628 dB, it still lags behind the PEPI-based final configuration. 
By creating a continuous panoramic epipolar plane, PEPI more effectively preserves global geometric structures, thereby providing higher-quality reconstruction. 
Notably, these three variants have identical parameter scales and computational costs, confirming that the benefits of PER arise solely from improved representational ability rather than from increased model capacity.

\subsubsection{Representation-Aware Asymmetric Scanning Strategy}
To validate the Representation-Aware Asymmetric Scanning (RAAS) strategy, we conduct a quantitative ablation on scanning configurations for the $4\times$ LFSR task, as summarized in Table \ref{table:scanning}. 
Exp. (a) serves as the symmetric-scanning baseline, employing quad-directional state updates following the standard SS2D design.
Exp. (b) applies RAAS exclusively to the SAI branch; pruning the inverse trajectories reduces FLOPs by 6.7\% with negligible performance impact. This indicates that inverse paths provide limited additional benefits in spatial domains with locally symmetric dependencies.
Exp. (c) extends RAAS to the EPI domain by pruning to a single epipolar-aligned traversal, while preserving full coupling in MacPI. This achieves a 13.5\% reduction in FLOPs relative to Exp. (a) without noticeable degradation. 
Conversely, Exp. (d) additionally prunes the MacPI paths, resulting in a clear drop in accuracy. This suggests that the MacPI domain necessitates comprehensive directional coupling to support complex mixed spatial-angular interactions, and that excessive pruning hinders effective cross-view information propagation.

\begin{table*}[!t]
\centering
\caption{Ablation study on different representation scanning strategies for 4$\times$ LFSR.}
\begin{tabular}{cccccccc}
\toprule[2pt]
 & SAI &  MacPI &  H-EPI & V-EPI  &  PSNR / SIMM &  Params & FLOPS\\
 \midrule[1pt]  
     (a) & $\Phi_{4\text{-}path}$  & $\Phi_{4\text{-}path}$ & $ \{\Phi_{row}, \Phi_{row}^{'}\}$ &  $ \{\Phi_{col}, \Phi_{col}^{'}\}$ &  32.703 / 0.9473 &   1.013 & 36.050 \\
    (b) & $\{\Phi_{row}, \Phi_{col}\}$  & $\Phi_{4\text{-}path}$ & $\{\Phi_{row}, \Phi_{row}^{'}\}$& $ \{\Phi_{col}, \Phi_{col}^{'}\}$&  32.700 / 0.9471 &   0.983 & 33.625 \\
    (c) & $\{\Phi_{row}, \Phi_{col}\}$  & $\Phi_{4\text{-}path}$ & $ \{\Phi_{col}\}$& $ \{\Phi_{col}\}$&  32.696 / 0.9471 &   0.953 & 31.200 \\
    (d) & $\{\Phi_{row}, \Phi_{col}\}$  &  $\{\Phi_{row}, \Phi_{col}\}$ & $ \{\Phi_{col}\}$& $ \{\Phi_{col}\}$&  32.329 / 0.9440 &   0.923 & 28.775 \\
\toprule[2pt]
\end{tabular}
\label{table:scanning}
\end{table*}

\begin{table}[!t]
\centering
\caption{Investigation on the scalability of the proposed RASLF with respect to network depth (number of PGR blocks $M$) and width (feature dimensions $C$).}
\begin{tabular}{ccccc}
\toprule[2pt]
  M&  C& PSNR / SIMM &  Params & FLOPS\\
 \midrule[1pt]  
  4  & 64 & 32.690 / 0.9472 &   0.953 &  31.200 \\
  \midrule[1pt]
  8  & 64 & 32.901 / 0.9492 &   1.654 &  55.756 \\
  12 & 64 & 32.972 / 0.9497 &   2.310 &  76.676 \\
  16 & 64 & 33.070 / 0.9502 &   2.996 & 100.020 \\
  \midrule[1pt]
  4 & 128 & 33.039 / 0.9498 &   3.486 & 101.185 \\
  4 & 256 & 33.129 / 0.9508 &  13.412 & 367.205 \\
\toprule[2pt]
\end{tabular}
\label{tab:scaling}
\end{table}

\subsection{Model Scaling Analysis}


To evaluate the scalability of RASLF, we investigated the impact of network depth (number of PGR blocks, $M$) and width (feature dimension, $C$) on the performance-efficiency trade-off.
As shown in Table \ref{tab:scaling}, increasing $M$ from 4 to 16 with a fixed $C=64$ consistently improves performance with linear growth in complexity, confirming the effectiveness of the DAA module in reducing redundancy in deep architectures.

Notably, a comparison between configurations with similar Flops, specifically the deeper variant ($M=16, C=64$) and the wider variant ($M=4, C=128$), shows that increasing cascading depth is more effective than expanding feature dimensionality. 
The former achieves higher accuracy with a more compact parameter footprint, providing empirical evidence that depth-wise cascading is more critical than width-wise expansion for efficient light-field modeling. 
This cascaded structure facilitates the continuous refinement of geometric representations through iterative state-space evolutions. 
Conversely, expanding the feature dimension to 256 leads to a disproportionate increase in parameters and FLOPs, with diminishing marginal utility. 
Therefore, to achieve the best balance between accuracy and computational cost, the configuration of $M=4$ and $C=64$ is chosen as the standard RASLF baseline.

\section{Conclusion}
This study presented RASLF, a representation-aware state space framework designed to facilitate deep collaboration among diverse LF representations
By integrating Progressive Geometric Refinement (PGR) with a Panoramic Epipolar Representation (PEPI), the proposed method successfully ensures global geometric consistency.
Furthermore, the Representation-Aware Asymmetric Scanning (RAAS) strategy and the Dual-Anchor Aggregation (DAA) module effectively eliminate directional and hierarchical redundancies, respectively, thereby improving the balance between reconstruction accuracy and computational efficiency.
Extensive evaluations on five public benchmarks confirmed that RASLF achieves state-of-the-art (SOTA) performance with remarkably compact parameters.

Future research will investigate the integration of light-field geometric priors with the intermediate states of the state-space model, promoting synergistic representations between SSM and LF features to further improve cross-view consistency and high-frequency texture restoration.


\bibliographystyle{IEEEtran}
\bibliography{bibtex/IEEEabrv, reference}

\end{document}